\documentclass[11pt,a4paper]{article}
\usepackage[hyperref]{acl2020}
\usepackage{times}
\usepackage{latexsym}

\usepackage{algorithm2e}
\usepackage{amsmath,amsfonts,amssymb,amsthm}
\usepackage{mathtools}
\usepackage{commath}

\usepackage{booktabs}

\usepackage{dcolumn}
\newcolumntype{z}[1]{D{.}{.}{#1}}

\usepackage{microtype}

\aclfinalcopy 

\title{Dual Past and Future for Neural Machine Translation}

\author{
Jianhao Yan\qquad Fandong Meng \qquad Jie Zhou\\
Pattern Recognition Center, WeChat AI, Tencent Inc, Beijing, China \\
elliottyan@tencent.com
}

\date{}

\begin{document}
\maketitle

\begin{abstract}
Though remarkable successes have been achieved by Neural Machine Translation (NMT) in recent years, 
it still suffers from the inadequate-translation problem. 
Previous studies \citep{zheng2018modeling,zheng2019dynamic} show that explicitly modeling the translated (\emph{Past}) and un-translated (\emph{Future}) contents of the source sentence is beneficial for translation performance. However, it is not clear whether the commonly used heuristic objective is good enough to guide the \emph{Past} and \emph{Future}. In this paper, we present a novel dual learning framework that leverages both source-to-target and target-to-source NMT models to provide a more direct and accurate supervision signal for the \emph{Past} and \emph{Future} modules. Experimental results demonstrate that our proposed method significantly improves the adequacy of NMT predictions and surpasses previous methods in two well-studied translation tasks.
\end{abstract}

\section{Introduction}

Neural Machine Translation (NMT) has achieved unprecedented successes and drawn much attention from both academia and industry.
Following the sequence-to-sequence learning paradigm, NMT approaches~\cite{sutskever2014sequence,bahdanau2014neural,vaswani2017attention,ott2018scaling} usually consist of two parts -- the encoder and the decoder, where the encoder maps the source side sentence into a sequence of hidden representations, and the decoder generates the target side tokens step by step based on the encoder outputs. 

Despite its success, the commonly used encoder-decoder framework in NMT always suffers from over- and under- translation problems \cite{tu2016modeling,tu2017neural}. The decoder may tend to repeatedly focus on same parts of the source sentence while ignoring the other parts.
Many efforts \cite{tu2016modeling,meng2018neural,zheng2018modeling,zheng2019dynamic} have been made to mitigate this issue 
by either explicitly or implicitly modeling the step-by-step translated and un-translated information during the decoding process. 
One promising direction is to track the translated (\emph{Past}) and un-translated (\emph{Future}) components of the source sentence \cite{zheng2018modeling,zheng2019dynamic} at each decoding step. The components are modeled by RNN or Capsule Network with heuristic objectives (e.g., Bag-of-Words Loss). 


\begin{figure}[t]
\centering
\includegraphics[scale=0.25]{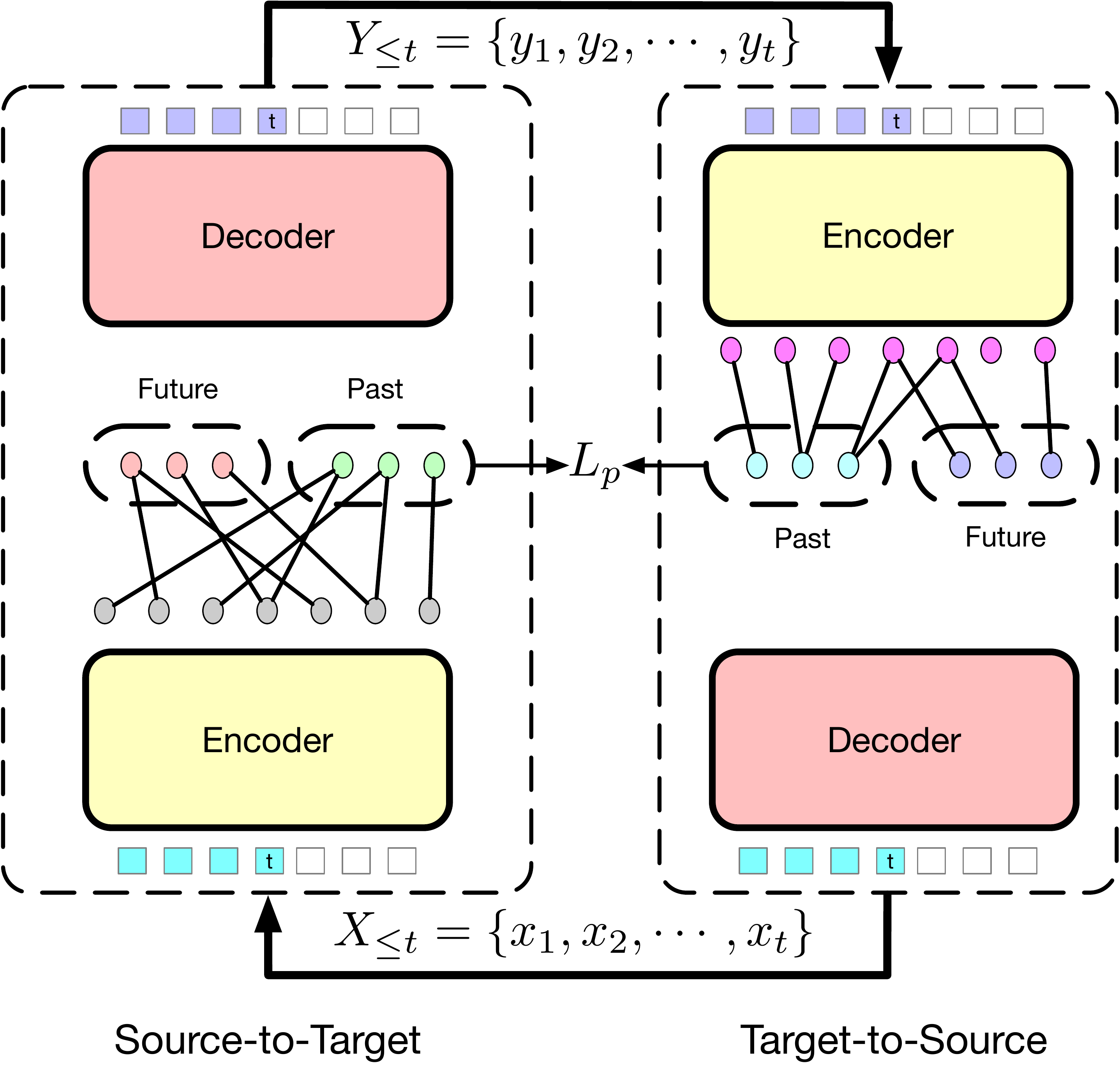}
\caption{The model architecture for Dual Past-Future-Transformer. Here we only depict the supervision of \emph{Past} module for simplicity. 
At decoding step $t$, the partial sub-sequence $X_{\leq t}$ and $Y_{\leq t}$ are fed into the reverse direction models and get the encoder outputs. Then, we use reverse directions' outputs from \emph{Past} to regularize the forward \emph{Past} module.} 
\label{overview}
\end{figure}


In this paper, we argue that the heuristic objectives in previous approaches may be indirect and insufficient in certain circumstances, which limits their effectiveness. 
The \emph{past} and \emph{future} modules have two major functionalities, which are the identification of past and future contents and extracting useful features for further predictions. 
However, prior studies mix these two functionalities up and try to model them jointly by only fitting the outputs of \emph{Past} / \emph{Future} module.
Here, we propose a novel dual learning method to enhance both two functionalities with two transformer models (source-to-target and target-to-source) trained simultaneously (See Figure \ref{overview}). 
On the one hand, we propose to use backward NMT encoder with the partially inputs to provide contextually-rich supervision for the past / future identification instead of a coarse-grained bag-of-word loss. 
On the other hand, we exploit a Guided Capsule Network \cite{zheng2019dynamic} on two encoders to align the capability of feature extraction with manually masking, instead of mixing up both functionalities. With the training proceeds, bidirectional models perform as teachers for each other and strengthen the performance iteratively.

We evaluate our approach on two commonly used translation datasets, i.e., the NIST Chinese-to-English task and the WMT 2014 English-to-German task. 
The experimental results demonstrate that our method significantly outperforms the previous strong baselines in terms of the translation quality of generated NMT translations. Also, among the subjective evaluation, our method surpasses previous adequacy-oriented methods in mitigating both over- and under-translation problem.

\section{Background}
Neural Machine Translation (NMT) often adopts an encoder-decoder framework. 
Given a source sentence $X=\{x_1, x_2, \cdots, x_I\}$ with $I$ words, and a target sentence $Y=\{y_1, y_2, \cdots, y_T\}$ with $T$ words, the encoder first maps source sentence $X$ into a sequence of word embeddings $E=\{e(x_1), \cdots, e(x_I)\}$. 
Then, it encodes the word embeddings into corresponding hidden representations $h=F(X)$ using its transformation layers. 
Similarly, the decoder follows the same procedure to encode the decoder inputs $\tilde{Y}$ (shifted target sequence $Y$ with a special $\langle \rm SOS \rangle$ token as head) into hidden representations $z$ in additional to taking $h$ into account. 

Then, the NMT model predicts the target sequence by maximizing the conditional probability based on $h$ and $\tilde{Y}_{<t}$:
\begin{equation}
  P(Y|X;\theta_{mt})=\prod_{t\in\{1,\cdots,T\}} P(y_{t}| \tilde{Y}_{<t}, h; \theta_{mt}),
\end{equation}
where $\theta_{mt}$ is the set of learnable parameters and $\tilde{Y}$ is a partial input. And, the commonly used training loss is the Cross Entropy Loss.
\begin{equation}
  L(\theta_{mt}) = \frac{1}{|D|} \sum_{(x,y) \in D} - \log P(Y|X; \theta_{mt}).
\end{equation}

\section{Proposed Method}

In this section, we introduce our dual learning framework for \emph{Past} and \emph{Future}, which not only explicitly models the dynamics of the translated (past) and un-translated (future) part of the source sentence, but also leverages the duality of source-to-target and target-to-source models to provide a more acurate and contextually-rich supervision.

More specifically, after the computation of encoder and decoder hidden states, we feed back the partially decoded sub-sequence for one model into the reverse model, and exploit a Guided Capsule Network to align the outputs for both directions' encoders outputs. Then, in a dual learning manner, two models are trained simultaneously and improve the translation performance. 

In the following section, we refer to one model as forward model and the other as backward model to illustrate our method in details.

\subsection{Guided Capsule Networks}
Firstly, we present the details of the Guided Capsule Network \cite{zheng2019dynamic}. 

Capsule Network \cite{sabour2017dynamic} has shown its superiority in solving the problem of assigning parts to wholes \cite{sabour2017dynamic}. In our settings, the capsule's routing by agreement mechanism is suitable in finding Past and Future in whole sentence \cite{zheng2019dynamic}. 

Formally, we regard the outputs from the forward encoding module $\{h_1, h_2, \cdots, h_I\}$ as the low-level capsules with a linear projection,
\begin{equation}
    u_{ij} = W_j * h_{i},
\end{equation}
where $W_{j} \in \mathbb{R}^{D*D_c}$ is a trainable transformation matrix for capsule $j$. 
Then, in the dynamic routing process, each vector representation for high-level capsule $\Omega_j$ is calculated by a \emph{squash} function,
\begin{align}
    \Omega_j &= {\rm squash}(s_j) = \frac{\norm{s_j}^2}{1 + \norm{s_j}^2}, \\
    s_j &= \sum_i^I c_{ij} u_{ij}, ~ c_{ij} = {\rm Softmax}(b_{ij}),
\end{align}
where $s_j$ is the weighted sum over all low-level capsules and $c_{ij}$ is the assignment probabilities (i.e., the agreement between low-level and high-level capsules). Note that, the high-level capsules are evenly split into two groups, representing the \emph{Past} $\Omega^P$ and \emph{Future} $\Omega^F$.

Then, during each iteration of Capsule Network, $b_{ij}$ is updated by a guided agreement between different level capsules with decoder output $z_t$,
\begin{equation}
    b_{ij} = b_{ij} + w^T {\rm tanh} (W_b [z_t;u_{ij};\Omega_j]).
\end{equation}

\subsection{Dual Past and Future}
After capturing the \emph{Past} and \emph{Future}, next we introduce how to use duality for supervising the \emph{Past} and \emph{Future}. In the following section, we refer to the dynamic guided capsule as \emph{DGC} for short.

Suppose in decoding time step $t$,  
the \emph{Past} capsule outputs for forward and backward models are,
\begin{align}
    \Omega^{P,b} &= {\rm DGC}(h^b, c_{ij}, z_t), \\
    \Omega^{P,f} &= {\rm DGC}(h^f, c_{ij}, z_t).
\end{align}
Then, we have a partially decoded target sequence $Y_t = \{y_1, \cdots, y_t\}$ from source-to-target and a partially decoded source sequence $X_t = \{x_1, \cdots, x_t\}$ from the other.
We put these partial outputs back into encoders of their reverse direction models, respectively:
\begin{align}
    \tilde{h^b_t} &= F^b(Y_{\leq t} = \{y_1, y_2, \cdots, y_t\}), \\
    \tilde{h^f_t} &= F^f(X_{\leq t} = \{x_1, x_2, \cdots, x_t\}),
\end{align}
where $\tilde{h^f_t}$ and $\tilde{h^b_t}$ denote the hidden states for partially decoded sub-sequence from both directions. Notably, $\tilde{h^f_t}$ and $\tilde{h^b_t}$ represent the contextual rich representation for translated words for either side. 

Then, we use another \emph{DGC} to extract the feature outputs $\tilde{\Omega}$ for $h^b_t$ and $h^f_t$:
\begin{align}
    \tilde{\Omega}^{P,b}_j &= {\rm DGC}(F^b(Y_{\leq t}), c_{ij} \cdot m_p), \\
    \tilde{\Omega}^{P,f}_j &= {\rm DGC}(F^f(X_{\leq t}), c_{ij} \cdot m_p),
\end{align}
where $m_p$ is the corresponding past mask for $t$-th step. 
After masking out the irrelevant assignment probability $c_{ij}$, the low-level capsules from translated words are only routed to the \emph{Past} capsules. 

Then, we can minimize the semantic distance between \emph{Past} capsules' outputs of both directions,
\begin{equation}
    L^P = \norm{\Omega^{P,f}_j - \tilde{\Omega}^{P,b}_j}_2 + \norm{\Omega^{P,b}_j - \tilde{\Omega}^{P,f}_j}_2
\end{equation}

Similarly, we perform the same computation for the \emph{Future} capsules except for feeding $X_{\geq t}$ and $Y_{\geq t}$. By this way, the bi-directional models are  improved in an iterative manner.\footnote{
Note that, for the consideration of computational efficiency, we actually put the whole sequence back into the encoder $F^b$ and set the attention bias to lower-triangle bias (past) with very large negative numbers (i.e., -1e9), the same as the bias used in the decoding self-attention process. }

\subsection{Incorporating with NMT}
The above approach can be applied on top of the general sequence-to-sequence model. In our experiments, we use Transformer \cite{vaswani2017attention} as our base model since it achieves many state-of-the-art results in the NMT task. 

Given the last layer encoder and decoder outputs, $h$ and $z_t$, we use \emph{DGC} to extract the \emph{Past} and \emph{Future} memory features from the source encoding side and obtain the holistic context for each decoding step. Following the setting in \citep{zheng2019dynamic}, an extra redundant capsule $\Omega^R$ is introduced,
\begin{align}
    &\Omega^P, \Omega^F ,\Omega^R = {\rm DGC} (z_t, h), \\
    &o_t = {\rm Linear}([z_t;\Omega^P;\Omega^F;\Omega^R]) + z_t,
\end{align}
where $\Omega^P$, $\Omega^F$ and $\Omega^R$ are the \emph{Past}, \emph{Future} and \emph{Redundant} capsule outputs, and [;] represents the concatenation operation. Finally, the output probability for each decoding step is computed via a softmax layer,
\begin{equation}
    P(y_t|y_{<t}, x) = {\rm Softmax}(o_t).
\end{equation}

\begin{table*}
  \centering
  \renewcommand{\arraystretch}{1.1}
  \setlength{\tabcolsep}{1.1mm}{
  \begin{tabular}{l | c | c || c ||c|c|c|c||c}
  \toprule
  \textbf{System} & \textbf{Speed} & \textbf{MT06} & \textbf{MT02} & \textbf{MT03} & \textbf{MT04} & \textbf{MT05} & \textbf{Average} & $\Delta$ \\ 
  \hline
  \multicolumn{8}{c}{\em Existing NMT systems} \\
  \hline
  \citet{wang2018switchout}  & - &45.47 & 46.31  & 45.30 & 46.45 & 45.62 & 45.83 & -\\
  \citet{cheng2018towards}  & - & 45.78  & 45.96 & 45.51 & 46.49 & 45.73 & 45.89 & -\\
  \hline
  \multicolumn{8}{c}{\em Our NMT systems} \\
  \hline
  \citet{vaswani2017attention}   & 1.00 &44.70&45.26&43.75 & 45.68  & 44.14  & 44.71 & reference \\
  \citet{zheng2019dynamic}  & 0.87 &45.70&46.13&44.90&46.84&45.20& 45.75 & +1.04   \\
  \hline
  Ours  & 0.87 & 45.96 & 46.29 & 44.83 & 46.92 & 45.26 & 45.85 & +1.14\\
     Ours + Inde.Train.  & 0.87 & \textbf{46.15} & \textbf{46.54} & \textbf{45.15} & \textbf{46.97} & \textbf{45.41} & \textbf{46.04} & \textbf{+1.33} \\

   \hline
  \end{tabular}}
  \caption{\label{tab:main} Case-insensitive BLEU scores (\%) on the NIST Chinese-to-English (ZH-EN) task. The improvements over the Transformer baseline~\cite{vaswani2017attention} are in the superscript.} 
  \vspace{-5pt}
\end{table*}
\section{Experiments}
\label{sec:exp}
\subsection{Dataset}
The main experiments are conducted in the widely used NIST Chinese to English (ZH-EN) dataset, containing 1.25M parallel sentences. We also show the results on WMT14 English to German (EN-DE) dataset, containing 4.50M parallel sentences, to compare our model performance with other state-of-the-art models. 

\subsection{Settings}
Among all of our experiments, we follow the \emph{Transformer-base} configuration from \cite{vaswani2017attention}. 
The residual dropout rates are 0.4 for NIST ZH-EN and 0.1 for WMT14 EN-DE. The dimension of \emph{Past} and \emph{Future} capsules is set to 256 and each component consists of 2 capsules. 
For NIST ZH-EN task and WMT14 EN-DE task, we use case-insensitive and case-sensitive 4-gram BLEU score

\begin{table}
  \centering
  \setlength{\tabcolsep}{0.3mm}{
  \begin{tabular}{l r r}
  \textbf{System} & \textbf{EN-DE}\\
  \hline
  \hline
  ConvS2S \cite{gehring2017convolutional} & 25.20 \\
  Transformer \cite{vaswani2017attention} & 27.30 \\

  \hline
  \hline
  Transformer (Our Impl.) & 27.54 \\
  \textbf{Ours} & \textbf{27.86} \\

  \end{tabular}}
  \captionsetup{justification=centering}
  \caption{\label{tab:wmt}Experimental results on WMT 2014 English-to-German (EN-DE) task.} 
  \vspace{-5pt}
\end{table}

\subsection{Main Results} 

We mainly conduct our experiments on the NIST Chinese-to-English (ZH-EN) translation task. 
Besides our implemented NMT systems, we also list the performance of several existing systems \cite{wang2018switchout,cheng2018towards} to support the effectiveness of our model.
The experimental results can be found in Table \ref{tab:main}. Compared with previous adequacy-oriented method \cite{zheng2019dynamic} and existing systems, our model shows its superiority with \textbf{+1.33} BLEU score improvement over the baseline Transformer. Although our model introduces approximately once more numbers of the parameters in Transformer-base, it does not hurt the decoding speed too much (0.87$\times$) because we use dual models for training but only one side model is used for testing. Also, we report the performance of EN-ZH translation produced by our model. Compared with Transformer baseline, our method improves the translation by \textbf{+0.66} BLEU scores.

\subsection{WMT14 EN-DE}
To evaluate our performance with other previous work, we also conduct experiments on the well-studied WMT 2014 English-to-German (EN-DE) translation task. The experimental results are shown in Table \ref{tab:wmt}.
Here we also list several results of previous state-of-the-art systems for comparison. We find that our model outperforms the previous results in the well-studied WMT-2014 EN-DE translation task.

\begin{table}
  \centering
  \setlength{\tabcolsep}{0.3mm}{
  \begin{tabular}{l | l | l }
  \textbf{System} & \textbf{Under (\%)} & \textbf{Over (\%)} \\
  \hline
  GDR \cite{zheng2019dynamic} & 71\% & 92\% \\
  Ours &  \textbf{82\%}$^{+11\%}$ & \textbf{95}\%$^{+3\%}$ \\
  \end{tabular}}
  \captionsetup{justification=centering}
  \caption{\label{tab:manual}Manually annotated results for under- and over-translation problem for generated predictions.} 
\end{table} 

\subsection{Subjective Evaluation}
Following prior work \citep{tu2016modeling,zheng2018modeling,zheng2019dynamic}, we also use the human evaluation to evaluate the adequacy of our proposed model. 
We randomly select 100 sentences from ZH-EN task and ask the annotator to judge whether the translations produced by GDR and our model suffer from under- or over- translation. The results demonstrate that our model outperforms in resolving both under- (+11\%) and over- (+3\%) translation cases, while there is still much room for improvement.

\section{Conclusion}
Sequence-to-sequence based neural machine translation models always suffer from the under- and over-translation problem. 
In this paper, we present a novel dual learning framework, aiming at modeling the translation adequacy. 
By leveraging the power of both source-to-target and target-to-source model, our proposed method provides a more direct and contextual-rich supervision signal for the translated and un-translated words. The experimental results demonstrate that our method outperforms the previous adequacy-based methods and achieves significant improvement in mitigating over- and under- translation problem.

\bibliography{anthology,acl2020}
\bibliographystyle{acl_natbib}

\end{document}